\documentclass{article}

\usepackage[main, final]{neurips_2026}
\usepackage[utf8]{inputenc} % allow utf-8 input
\usepackage[T1]{fontenc}    % use 8-bit T1 fonts
\usepackage{hyperref}       % hyperlinks
\usepackage{url}            % simple URL typesetting
\usepackage{booktabs}       % professional-quality tables
\usepackage{amsfonts}       % blackboard math symbols
\usepackage{nicefrac}       % compact symbols for 1/2, etc.
\usepackage{microtype}      % microtypography
\usepackage{xcolor}         % colors

\usepackage{placeins}
\usepackage{subcaption}
\usepackage{amsmath}
\usepackage{amssymb}
\usepackage{mathtools}
\usepackage{amsthm}
\usepackage{bm}
\usepackage{xspace}
\usepackage{wrapfig}

\newcommand{\method}{\textsc{NeuNeu}\xspace}

\hypersetup{
    colorlinks=true,
    citecolor=[rgb]{0.0, 0.3, 0.7},
    urlcolor=[rgb]{0.0, 0.3, 0.7},
    linkcolor=[rgb]{0.0, 0.3, 0.7},
}

\title{Neural Neural Scaling Laws}

\author{
Michael Y. Hu \quad Jane Pan \quad Ayush Rajesh Jhaveri \quad Nicholas Lourie \quad Kyunghyun Cho \\
New York University \\
\texttt{\{michael.hu,kyunghyun.cho\}@nyu.edu}
}

\begin{document}

\maketitle

\begin{abstract}

Neural scaling laws predict how language model performance improves with increased training inputs. While aggregate metrics like validation loss can follow smooth power-law curves, individual downstream tasks exhibit diverse scaling behaviors: some improve monotonically, others plateau, and some even degrade with scale. We argue that predicting downstream performance from validation loss suffers from two limitations: averaging token-level losses obscures signal, and no simple parametric family can capture the full spectrum of scaling behaviors.
To address this, we propose Neural Neural Scaling Laws (\method), a neural network that frames scaling law prediction as time-series extrapolation. \method~combines temporal context from observed accuracy trajectories with token-level validation losses, learning to predict future performance without the limitations inherent in assuming a specific functional form. Trained entirely on open-source model checkpoints from HuggingFace, \method~achieves 1.99\% mean absolute error in predicting model accuracy on 66 downstream tasks---a 44\% reduction compared to logistic scaling laws (3.56\% MAE). Furthermore, \method~generalizes zero-shot to unseen model families, architectures, parameter counts, and downstream tasks. Our work suggests that predicting downstream scaling directly from data outperforms parametric alternatives.

\end{abstract}

\section{Introduction}

Neural scaling laws characterize how language model performance improves with increased compute, data, and parameters \citep{kaplan2020scaling, hoffmann2022training}. These laws typically take the form of power-law relationships such as $L(C) = \alpha C^{-\beta}$, where $L$ is the reducible loss and $C$ is an input to training, like compute. Such simple functional forms have proven remarkably useful for predicting training dynamics and optimizing resource allocation.

However, the translation from pretraining loss to downstream performance is far more complex. While aggregate metrics like pretraining loss or task performance averaged over many domains follow smooth scaling curves, individual tasks exhibit diverse behaviors as they scale: some improve monotonically, others plateau, and some even degrade with scale---a phenomenon known as inverse scaling \citep{mckenzie2023inverse}. Taken together, it seems that no single parametric family can capture the full spectrum of scaling behaviors \citep{lourie-etal-2025-scaling}.

We hypothesize that predicting future performance from validation loss suffers from two flaws, both of which limit the usefulness of downstream scaling laws: first, validation loss creates a bottleneck by averaging rich token losses into a single, obscured signal; and second, no simple hypothesis class exists to fit all behaviors of downstream tasks. To fix these issues, we propose to use a neural network that predicts downstream task performance while incorporating token-level loss information.

\begin{figure*}[t!]
    \centering
    \vspace{0.5em}% <-- top margin
    \makebox[\textwidth][c]{%
        \hspace{1em}% <-- left margin
        \begin{subfigure}[b]{0.62\textwidth}
            \centering
            \includegraphics[width=\textwidth]{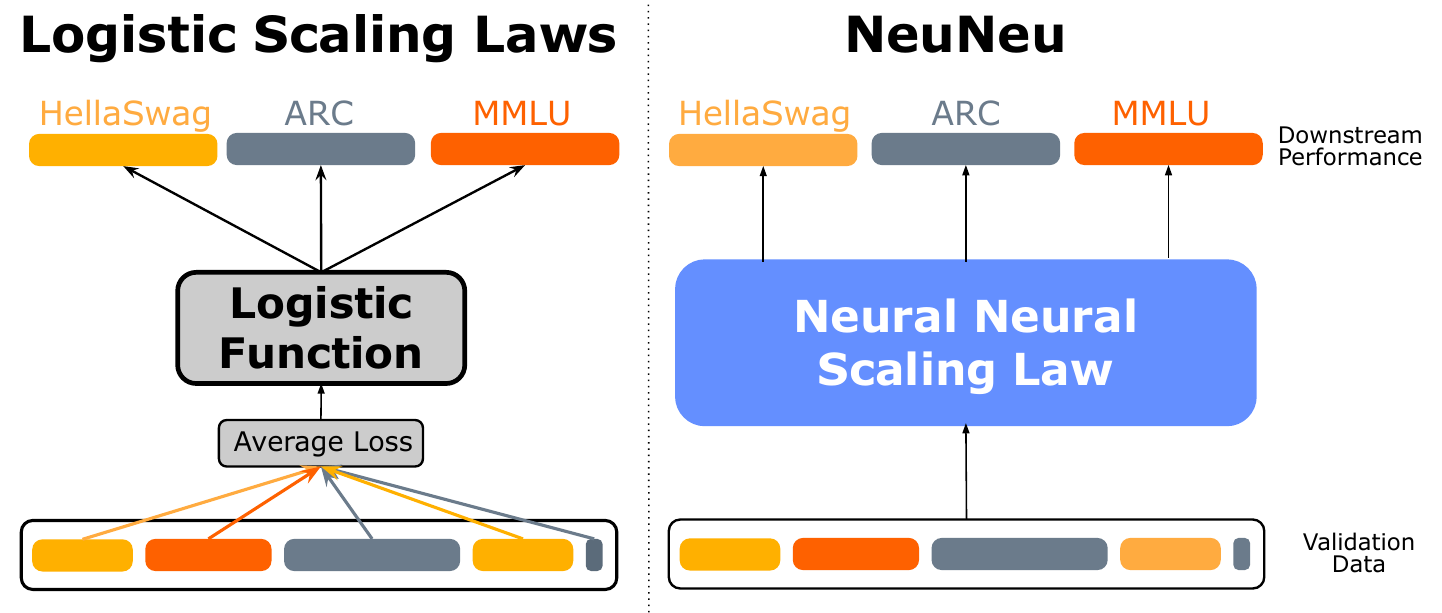}
        \end{subfigure}
        \hfill
        \begin{subfigure}[b]{0.34\textwidth}
            \centering
            \includegraphics[width=\textwidth]{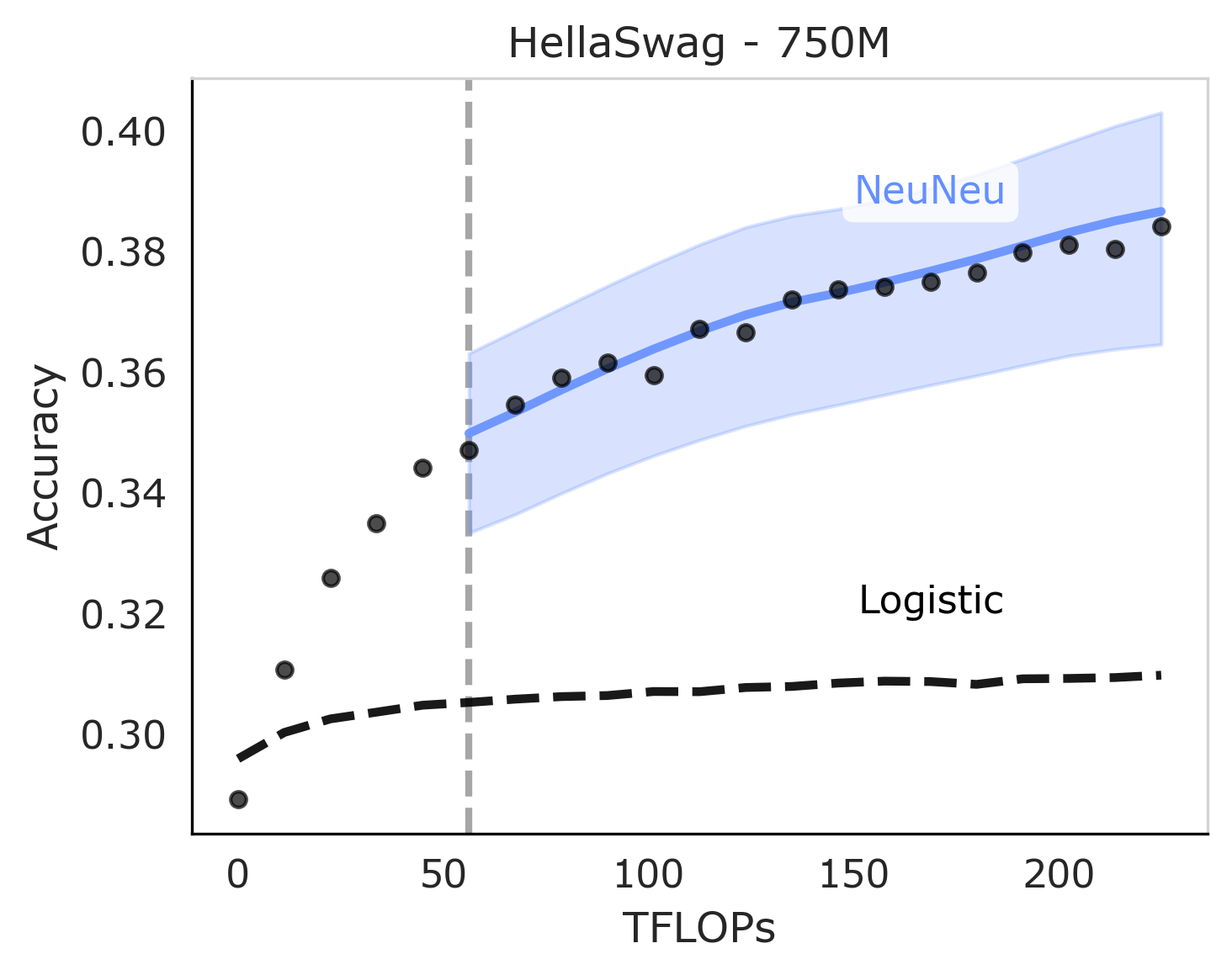}
        \end{subfigure}
        \hspace{1em}% <-- right margin
    }
    \vspace{0.2em}% <-- bottom margin (space before caption)
    \caption{
        Richer signal from token-level losses (center) enables \method~to better forecast accuracies for downstream tasks (right). Average validation loss, used in logistic scaling laws, averages away token-level loss changes.
    }
    \label{fig:conceptual-map}
\end{figure*}

Our ``neural'' neural scaling law, or \method, frames scaling law prediction as a time-series extrapolation problem.
Unlike parametric approaches that rely on aggregate metrics, \method~predicts downstream performance by combining observed accuracy trajectories with token-level validation losses. 
This allows the model to leverage the signal within loss distributions that averaging typically obscures, as well as the trends within accuracy trajectories that pointwise loss-to-accuracy mappings ignore.
To ensure generalization across unseen model families and parameter counts, we design inputs to be invariant across model scales. 
We achieve this by abstracting training steps into relative compute intervals and converting unbounded losses into token probabilities, enabling the network to learn patterns in training dynamics decoupled from the specific training configuration.

We train \method~on open-source language model training trajectories \citep{magnusson2025datadecide} from HuggingFace \citep{wolf-etal-2020-transformers}, enabling anyone to fit their own neural neural scaling laws without first performing a large number of training runs. \method performs time-series forecasting using token-level loss information and observed downstream accuracies. During evaluation, we use \method to forecast the remaining LM accuracy trajectories for all downstream tasks after observing the initial trajectories of validation losses and downstream accuracies.

Our results show that \method~achieves 1.99\% mean absolute error (MAE) on 66 downstream tasks, a 44\% reduction compared to the widely used logistic scaling laws \citep{magnusson2025datadecide,gadre2025language}. Furthermore, \method~is a more robust decision-making tool: it generalizes zero-shot to unseen tasks with significantly lower error, and correctly ranks the final performance of competing model configurations with 76.6\% accuracy, a 20.2\% relative improvement over logistic scaling laws. Our contributions:
\begin{enumerate}
    \item We propose \method, the first model that predicts downstream scaling performance without parametric or prior assumptions. On average, \method~outperforms logistic scaling laws by 44\% at predicting downstream accuracies across 66 tasks (\S \ref{sec:results}).
    \item We train \method~with quantile regression \citep{koenker2001quantile}, allowing the model to predict uncertainty. We find that \method's 10\%-90\% interquantile interval captures 77.6\% of the true data out of an expected 80\%, suggesting that \method produces calibrated uncertainty estimates (\S \ref{sec:calibration}).
    \item \method strictly outperforms baselines that use average validation loss \citep{LC-PFN,caballero2023broken}, demonstrating that average validation loss discards information usable by a more powerful model. As the corpus of open-source models and training trajectories grows, we advocate for more expressive scaling laws that scale with data (\S \ref{sec:discussion}).
\end{enumerate}

Code: \url{https://github.com/michahu/neuneu}

\begin{figure*}[t!]
    \centering
    \includegraphics[width=0.7\textwidth]{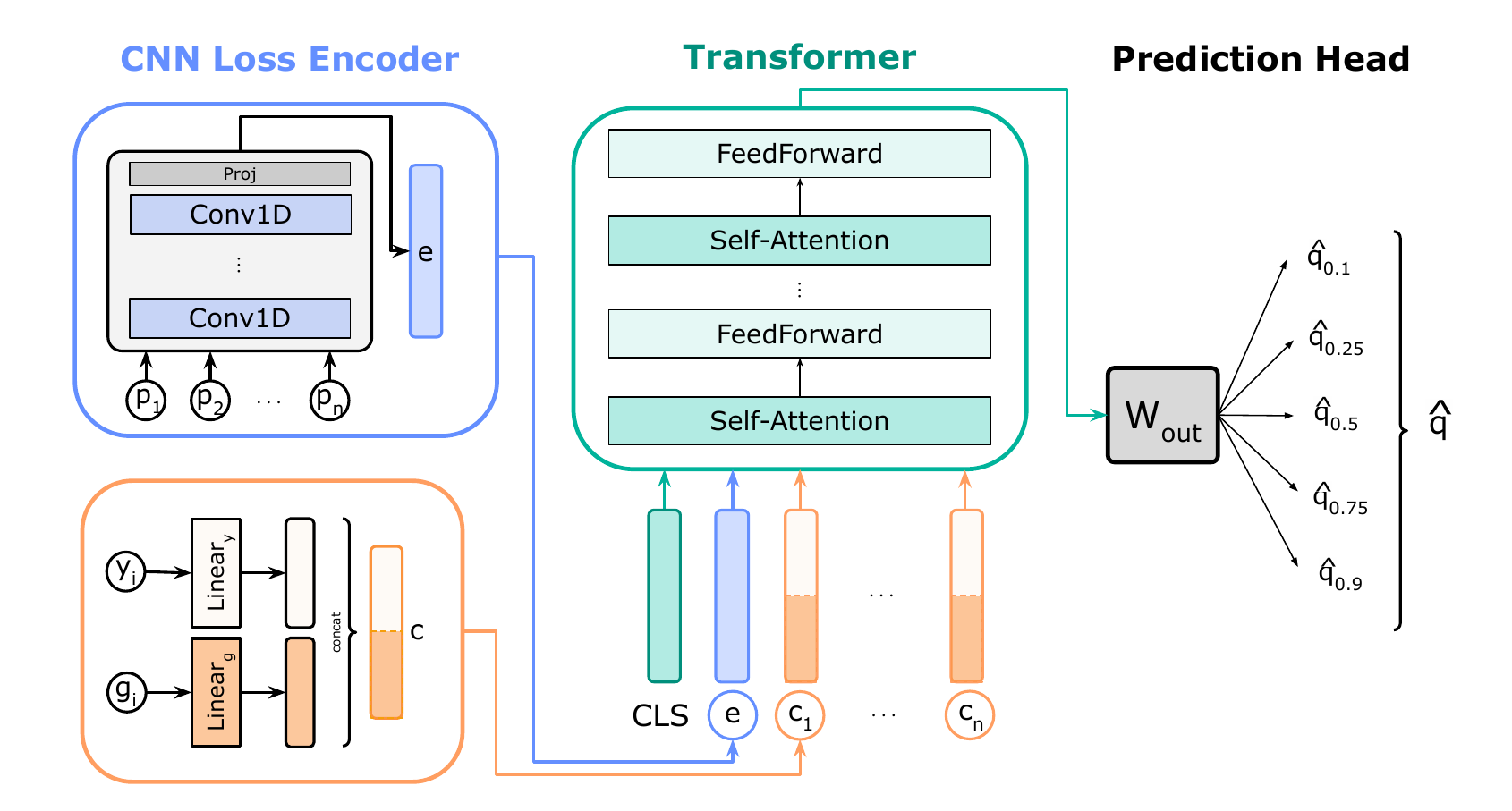}
    \caption{\method~encodes and processes token-level validation probabilities alongside a sequence of historical downstream accuracies and compute gaps, which are projected into context tokens. The BERT-style transformer \citep{devlin-etal-2019-bert} backbone uses this information to predict a distribution over the downstream accuracy via quantile regression on the [CLS] token representation.}
    \label{fig:architecture}
\end{figure*}

\section{Neural Neural Scaling Laws}

\paragraph{Problem setting.} This work focuses on improving downstream performance prediction from validation loss, the main metric during language model (LM) pretraining. Periodically, we evaluate a language model on a validation set of $N$ tokens and $D$ downstream tasks. Suppose we are at time $t$ and want to predict a language model's task performance at time $t+K$. We have the following information at our disposal:
\begin{itemize}
    \item Token-level loss vectors $\bm{\ell}_{1:t} \in \mathbb{R}^{t \times N}$
    \item Downstream accuracies $\mathbf{y}_{1:t} \in [0,1]^{t \times D}$
\end{itemize}

Logistic scaling laws solve this prediction problem by assuming that 1) the average validation loss $\bar{\ell}_t$ is sufficient to predict all downstream task performances $\mathbf{y}_t$ and 2) the relationship between validation loss and downstream task performance is well-described by a logistic function, which is reasonable for predicting values that transition between chance and a saturating threshold. One then fits the parameters of the logistic function:
\begin{align*}
    y_t^{(j)} &= f(\bar{\ell}_t; a, k, L_0, b) = \frac{a}{1+e^{-k(\bar{\ell}_t - L_0)}} + b
\end{align*}

These scaling laws have high bias and low expressivity. In moving to a neural network, which is more expressive, we must choose input and output representations that allow the neural network to extrapolate. In \S\ref{sec:representation}, we discuss our loss representation choices, present the architecture in \S\ref{sec:architecture}, and finish with training and evaluation details in \S\ref{sec:data} and \S\ref{sec:evaluation}.

\subsection{Validation Loss Representations}
\label{sec:representation}

One drawback of logistic scaling laws is that the average loss $\bar{\ell}_t$ does not retain distributional information, which we hypothesize is beneficial for predicting downstream performance. Two models can achieve the same validation loss with loss distributions of different skews or variances, which could be indicative of different underlying capabilities.

First, to fix the issue with cross-entropy loss being unbounded, we convert token-wise losses into token-wise probabilities:
\begin{equation*}
    p_{t,i} = e^{-\ell_{t,i}} \quad \text{for } i = 1, \ldots, N
\end{equation*}
In general, we found that training on probabilities leads to better neural models than training on losses; see Figure \ref{fig:prob-ablation} for discussion and ablations. To test our hypothesis about distributional information, we consider three neural models, distinguished by how they use token probabilities:
\begin{itemize}
    \item \method: Takes the token probability vector $\mathbf{p}_t \in \mathbb{R}^N$, where $N$ denotes the validation set size or a subsample thereof. We use $N=256{,}000$ in practice, roughly the size of one LM training minibatch.
    \item \textsc{Average}: Takes all observed average validation probabilities $\mathbf{\bar{p}}_{\leq t}$, similar to how logistic scaling laws use the average loss.
    \item \textsc{NoLoss}: Takes no information about token probabilities or losses. The model makes predictions using the downstream accuracies $\mathbf{y}_{1:t}$ only.
\end{itemize}

\subsection{Learning a Representation for Token Probabilities}
\label{sec:architecture}

Our three neural predictors observe a sequence of accuracies and compute gaps $y_1, g_1, \ldots, y_t, g_t$ and use this information to predict the next accuracy $y_{t+g_t}$ in the sequence. Gaps are units of compute between evaluation steps. For example, the sequence $(0.5, 1, 0.6)$ means that the LM's accuracy was 0.5, and after one unit of training compute (\emph{e.g.}, 500 TFLOPs), the accuracy is now 0.6. Abstracting compute as gaps induces invariance across LM training scales, enabling generalization.

Our neural predictors consist of three components: loss encoder, transformer, and prediction head. See Figure~\ref{fig:architecture}. We show the forward pass for \method first, then explain each component.
\begin{align}
\mathbf{e} &= \text{LossEncoder}(\mathbf{p}_{t}) \\
\mathbf{c}_i &= \text{concat}\big(\text{Linear}_y(y_i),\ \text{Linear}_g(g_i)\big) \quad \text{for } i = 1, \ldots, t \\
\mathbf{H} &= \text{Transformer}([\text{CLS}; \mathbf{e}; \mathbf{c}_1; \ldots; \mathbf{c}_t]) \\
\hat{\mathbf{q}} &= \mathbf{W}_{\text{out}} \cdot \mathbf{H}_{0}
\end{align}

At inference time, suppose the observed sequence is $(y_1{=}0.5, g_1{=}1, y_2{=}0.6)$ and we want to predict the accuracy 5 units of compute (\emph{e.g.}, 2,500 TFLOPs) into the future. We feed the token-level probabilities $\mathbf{p}_{t}$ to Equation (1) and the sequence $(y_1{=}0.5, g_1{=}1, y_2{=}0.6, g_2{=}5)$ to Equation (2), where $g_2{=}5$ is the query gap, similar to the approach in \citet{Che2018}. We then run the forward passes in Equations (3) and (4).

\paragraph{Loss Encoder.}
The loss encoder in Equation (1) produces an embedding $\mathbf{e}$. To process $N$ token probabilities, we use $L$ layers of learned 1D convolutions for hierarchical downsampling:
\begin{align*}
\mathbf{x}_0 &= \mathbf{p}_{t} \\
\mathbf{x}_i &= \text{GELU}(\text{GroupNorm}(\text{Conv1D}(\mathbf{x}_{i-1}))) \\ 
\mathbf{e} &= \mathbf{W}_{\text{proj}} \cdot \text{flatten}(\mathbf{x}_L)
\end{align*}
% \mathbf{p}_{t}$ contains the token probabilities and
where $\mathbf{W}_{\text{proj}}$ projects the flattened features to the hidden dimension of the transformer. We use $L=4$ Conv1D layers with kernel size $k=64$, stride $s=16$, and channels increasing through $(8, 16, 32, 64)$. The \textsc{Average} encoder projects each probability and averages to produce one embedding $\mathbf{e} = \frac{1}{t} \sum_{k=1}^{t} \text{Linear}(\bar{p}_k)$, and \textsc{NoLoss} has no loss encoder, omitting $\mathbf{e}$.

\paragraph{Transformer.}
The main sequence model is a standard transformer encoder with bidirectional attention \citep{attn,devlin-etal-2019-bert} and rotary embeddings (RoPE, \citet{rope}).
Recall from Equation (2) that each context element $\mathbf{c}_i \in \mathbb{R}^d$ is the concatenation of projected accuracy and gap values. The first half of $\mathbf{c}_i$ contains information about $y_i$, and the second half about $g_i$.

The input sequence to the transformer is $[\text{CLS}; \mathbf{e}; \mathbf{c}_1; \ldots; \mathbf{c}_t]$. The transformer processes this with 6 layers of pre-norm self-attention, and the output is predicted from the CLS token position.

\paragraph{Prediction Head.} Unlike logistic scaling laws, our model also captures prediction uncertainty via quantile regression \citep{koenker2001quantile}. The output embedding $\mathbf{H}_0$ from the CLS position is projected to $Q=5$ quantile predictions:
$\hat{\mathbf{q}} = [\hat{q}_{0.1}, \hat{q}_{0.25}, \hat{q}_{0.5}, \hat{q}_{0.75}, \hat{q}_{0.9}] = \mathbf{W}_{\text{out}} \cdot \mathbf{H}_{0}$.

Training uses pinball loss summed across quantiles $\mathcal{T} = \{0.1, 0.25, 0.5, 0.75, 0.9\}$. For a target accuracy $y$ and predicted quantile $\hat{q}_\tau$:
\begin{equation*}
\mathcal{L}_{\text{pinball}} = \sum_{\tau \in \mathcal{T}} \begin{cases}
\tau (y - \hat{q}_\tau) & \text{if } y \geq \hat{q}_\tau \\
(1-\tau) (\hat{q}_\tau - y) & \text{otherwise}
\end{cases}
\end{equation*}

The output $\hat{\mathbf{q}}$ provides a calibrated distribution over predicted accuracy at the target compute scale. The median $\hat{q}_{0.5}$ serves as the point estimate, while the interquantile interval $\hat{q}_{0.9} - \hat{q}_{0.1}$ captures uncertainty.

\paragraph{Inference efficiency.} \method contains 19.5M parameters, spread over 4 CNN and 6 transformer layers, finishing inference in a few seconds on an M-series Apple CPU. When collecting the input $\mathbf{p}_t$ for \method, the only difference from recording average validation loss is that one simply skips the averaging step; thus, the overhead from running \method to extrapolate scaling is minimal.

\subsection{Training Data}
\label{sec:data}

A critical step in training such a neural predictor is obtaining diverse LM pretraining trajectories. In this work, we demonstrate that such data is \textbf{freely available on HuggingFace}: anyone can train our model from open-source data. In particular, we train \method~using training runs of 6 model sizes from the DataDecide model suite \citep{magnusson2025datadecide}. Each training trajectory contains three random seeds, a variable number of checkpoints, and each checkpoint is evaluated on the 66 downstream tasks including the OLMES evaluation suite \citep{gu-etal-2025-olmes}. We use random seed 0 from the \{90M, 150M, 300M, 530M, 750M, 1B\} parameter models, saving other seeds for evaluation.

In total, we use 6 model sizes trained on 24 different pretraining datasets, yielding 144 unique pretraining trajectories, or $144 \times 66 = 9{,}504$ unique accuracy trajectories. For all training and evaluation details, see Appendix \ref{sec:reproducibility}.

\paragraph{Data augmentation.} For each model and task, we construct training samples from checkpoint accuracies as follows. Let $(y_1, y_2, \ldots, y_T)$ denote the sequence of accuracies at consecutive checkpoints. We first impute unit gaps to form the context sequence: $\mathcal{S}_0 = [(y_1, 1), (y_2, 1), \ldots, (y_T, 1)]$. 

To create multiple examples from one training trajectory and a model that is robust to missing data, we randomly drop tuples from the sequence with probability $p=0.4$, as inspired by \citet{Che2018}. When tuple $i$ is dropped, its gap is absorbed into the preceding tuple:
\begin{align*}
[(y_{i-1}, g_{i-1}), (y_i, g_i), (y_{i+1}, g_{i+1})] \xrightarrow{\text{drop } y_i} [(y_{i-1}, g_{i-1} + g_i), (y_{i+1}, g_{i+1})]
\end{align*}

For a subsequence $\mathcal{S} = [(y_{s_1}, g_{s_1}), \ldots, (y_{s_k}, 1)]$ ending at checkpoint $s_k$, we generate training targets for all future checkpoints $j, s_k < j \leq T$. Let $g_{\text{target}} = j - s_k$. We simply replace the final $1$ with $g_{\text{target}}$ to create $\mathcal{S}'$, then add the appropriate input representation:
\begin{align*}
(\mathcal{S}', \mathbf{p}_{t}) &\mapsto y_j \quad (\method), \quad\quad\quad
(\mathcal{S}', [\bar{p}_1, \ldots, \bar{p}_{s_k}]) \mapsto y_j \quad (\textsc{Average}) 
\end{align*}

We evaluate all checkpoints on a shard of the WebOrganizer dataset \citep{wettig2025organize} and save the token probabilities. When training \method, we sample random spans of length 256{,}000. To generalize across tokenizers from different models, we use whitespace tokenization and combine probabilities of subwords, following \citet{tjuatja-neubig-2025-behaviorbox}. See Table \ref{tab:hyperparameters} for hyperparameters.

\subsection{Evaluation and Baselines}
\label{sec:evaluation}

We test generalization of all scaling laws on four kinds of unseen language model pretraining runs: new 1) random seeds, 2) pretraining datasets, 3) model families, and 4) downstream tasks. We use:
\begin{itemize}
    \item Random seed 2 from DCLM-Baseline training runs with parameter sizes \{90M, 150M, 300M, 530M, 750M, 1B\} \citep{magnusson2025datadecide}.
    \item C4 training runs with parameter sizes \{90M, 150M, 300M, 530M, 750M, 1B\}, seed 0. We withhold all C4 runs from the scaling laws' training data \citep{magnusson2025datadecide}.
    \item Pythia \citep{pythia} runs with parameter sizes \{70M, 1.4B, 2.8B, 6.9B, 12B\} and OLMo-Hybrid-7B \citep{merrill2026olmohybrid}. These model sizes lie outside of the training data distribution of our predictive models and represent challenging shifts in pretraining dataset, architecture, and parameter size. In particular, OLMo-Hybrid-7B is not a transformer---it mixes attention and recurrent neural network layers.
    \item All three conditions above while also withholding 13 randomly selected tasks of the 66 in DataDecide during training. This tests \method's ability to \textbf{generalize zero-shot to unseen tasks. This is impossible with logistic scaling laws}, which fit a separate model per task.
\end{itemize}

\paragraph{Baselines.} We train all predictors on the training data described in \S \ref{sec:data}. At inference time, we condition all transformer models on accuracies from the first 20\% of each heldout training run and compute mean absolute error (MAE) for all methods on the unobserved 80\%.
\begin{itemize}
    \item \textsc{Logistic:} Logistic scaling law, fitted per task. To give \textsc{Logistic} the best possible chance, we feed it ground truth average loss $\bar{\ell}_{t+K}$ for checkpoint $t+K$: $\hat{y}_{t+K}^{(i)} = f(\bar{\ell}_{t+K}; a, k, L_0, b)$. We assume that whatever prediction method one uses to obtain $\bar{\ell}_{t+K}$ is perfect, and focus on the problem of downstream prediction. Thus, logistic scaling laws have \textit{a strict advantage}.
    \item \textsc{LC-PFN:} A meta-learned, open-source transformer model that performs in-context Bayesian inference over learning curves \citep{LC-PFN}.
    \item \textsc{BNSL:} Broken neural scaling laws \citep{caballero2023broken}, which relax the assumptions of logistic scaling laws by allowing a break or transition between different scaling regimes. Also receives $\bar{\ell}_{t+K}$.
    \item \textsc{NoLoss, Average, NeuNeu}: Discussed in \S \ref{sec:representation} and \S \ref{sec:architecture}. If our hypothesis that token-level information is useful holds, then \method should outperform \textsc{NoLoss} and \textsc{Average}.
\end{itemize}

For our neural models, we report standard deviation $\pm2\sigma$ over 5 random seeds as an error bar. Randomness only impacts \textsc{Logistic} and \textsc{BNSL} in numerical imprecision, and \textsc{LC-PFN} only has one released random seed.

To reiterate, \textsc{Logistic} and \textsc{BNSL} receive the ground truth average loss at each timestep during inference, and they are fitted to entire trajectories during training. To do forecasting with these models, one typically fits another scaling laws to predict loss from compute \citep{gadre2025language,magnusson2025datadecide}. We decided not to do this because it introduces a second source of error. Instead, we simply gave \textsc{Logistic} and \textsc{BNSL} the benefit of perfect loss prediction, which \textsc{LC-PFN} and \method do not have.

\begin{figure*}[t]
\centering
\begin{subfigure}[b]{\textwidth}
    \centering
    \includegraphics[width=\textwidth]{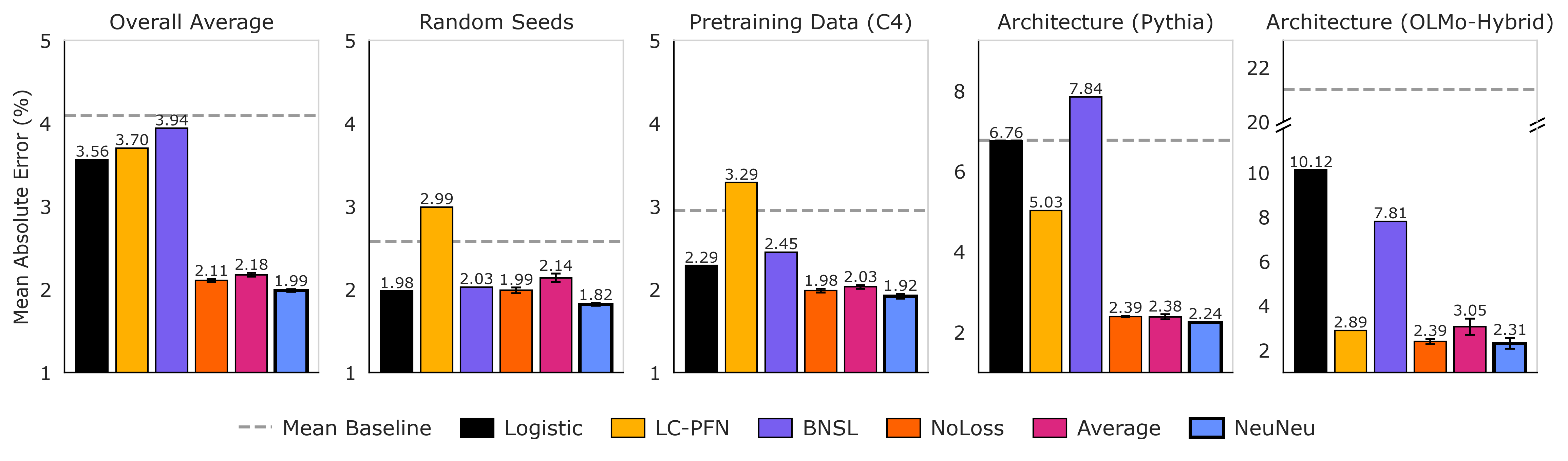}
    \caption{\method~significantly outperforms all other scaling laws at generalizing to new random seeds, pretraining data, and unseen transformer and non-transformer architectures. The gray dashed line denotes mean absolute error (MAE) of always predicting the average accuracy from the training set for each task.}
    \label{fig:main_plot}
\end{subfigure}

\vspace{1em}

\begin{subfigure}[b]{\textwidth}
  \centering
  \begin{tabular}{l|ccccccc}
      \toprule
      \textbf{Task} & Mean & Logistic & LC-PFN & BNSL & NoLoss & Average & \method \\
      \midrule
      ARC-Challenge & 0.0529 & 0.0320 & 0.0237 & 0.0331 & 0.0157 & 0.0220 & \textbf{0.0141} \\
      ARC-Easy      & 0.0787 & 0.0496 & 0.0253 & 0.0650 & 0.0218 & 0.0362 & \textbf{0.0191} \\
      BoolQ         & 0.0703 & 0.0647 & 0.0644 & \textbf{0.0518} & 0.0569 & 0.0542 & 0.0528 \\
      CSQA          & 0.0883 & 0.0468 & 0.0450 & 0.0518 & 0.0322 & 0.0463 & \textbf{0.0259} \\
      HellaSwag     & 0.0659 & 0.0401 & 0.0175 & 0.0291 & 0.0092 & 0.0179 & \textbf{0.0051} \\
      OpenBookQA    & 0.0553 & 0.0279 & 0.0198 & 0.0333 & 0.0146 & 0.0168 & \textbf{0.0135} \\
      PIQA          & 0.0456 & 0.0261 & 0.0164 & 0.0198 & 0.0138 & 0.0227 & \textbf{0.0074} \\
      SocialIQA     & 0.0237 & 0.0164 & 0.0178 & 0.0101 & 0.0090 & 0.0140 & \textbf{0.0081} \\
      WinoGrande    & 0.0345 & 0.0166 & 0.0383 & 0.0185 & 0.0177 & 0.0261 & \textbf{0.0146} \\
      MMLU          & 0.0387 & 0.0357 & 0.0381 & 0.0400 & 0.0211 & 0.0208 & \textbf{0.0202} \\
      \bottomrule
  \end{tabular}
  \caption{Mean absolute error for scaling law prediction on the OLMES tasks, averaged over all settings.}
  \label{tab:mae_results}
\end{subfigure}
\caption{Generalization results for downstream task accuracy prediction.}
\label{fig:main}
\end{figure*}

\section{Results}
\label{sec:results}

Having described our model and training setup, we now evaluate \method against parametric and neural baselines. We discuss \method's raw performance in \S\ref{sec:performance}, uncertainty calibration in \S\ref{sec:calibration}, and whether its predictions translate into better decisions about which models to train in \S\ref{sec:ranking}.

\subsection{\method Performs Well on All Evaluation Tasks}
\label{sec:performance}

In Figure \ref{fig:main_plot}, \method~achieves the lowest mean absolute error across all evaluation conditions. Table \ref{tab:mae_results} reports MAE for each OLMES task \citep{gu-etal-2025-olmes}, with MMLU tasks aggregated. 
\method is best on all tasks except BoolQ, where it is second-best.
Appendix Figure \ref{fig:horizon}A shows mean absolute error (MAE) on the 1B training runs from our evaluation set, and neural methods have lower MAE than logistic scaling laws and other baselines for every extrapolation horizon.
\method also outperforms the \textsc{NoLoss} and \textsc{Average} ablations, supporting our hypothesis that average validation loss discards distributional information helpful for downstream prediction.

\paragraph{Parametric scaling laws fail catastrophically on new architectures.}
Logistic and broken neural scaling laws perform progressively worse on more challenging generalizations. On Pythia and OLMo-Hybrid runs---which differ from the training trajectories in pretraining corpus, parameter count, and architecture---\textsc{Logistic} and \textsc{BNSL} incur around 4 times higher error than on in-distribution (different random seed) evaluations. This suggests that parametric scaling laws like \textsc{Logistic} and \textsc{BNSL} generalize poorly to large changes in training settings.

All methods outperform the trivial baseline of predicting average accuracy from the training set for each task (Figure \ref{fig:main_plot}, dashed lines). LC-PFN performs roughly on par with the logistic scaling laws, and worse than our neural methods; however, its predictions are not dramatically worse on Pythia and OLMo-Hybrid.
In Appendix Figure \ref{fig:horizon}B, we find that LC-PFN also improves when it sees more of the training curve before starting predictions, indicating that it is inferring properly.
Ultimately, \method has the advantage of being trained specifically to extrapolate language model scaling, whereas LC-PFN is trained to predict learning curves in general.
Overall, \method outperforms all baselines with non-overlapping error bars, the gap widening on the most challenging out-of-distribution evaluations.

\paragraph{\method~generalizes zero-shot to unseen downstream tasks.}
A key advantage of \method~over task-specific parametric fits like logistic or broken neural scaling laws is that a single trained model predicts accuracy for any task, including tasks never seen during meta-training. In Figure \ref{fig:task_and_quantile}A, we examine \method's performance after holding out 13 of the 66 OLMES tasks during training. MAE actually \emph{decreases} slightly on the unseen tasks, and---critically---remains lower than the MAE achieved by logistic scaling laws fit directly to those same tasks. In other words, our neural model predicts unseen tasks better than the parametric scaling laws trained on them.

\begin{figure*}[t]
\centering
\includegraphics[width=\textwidth]{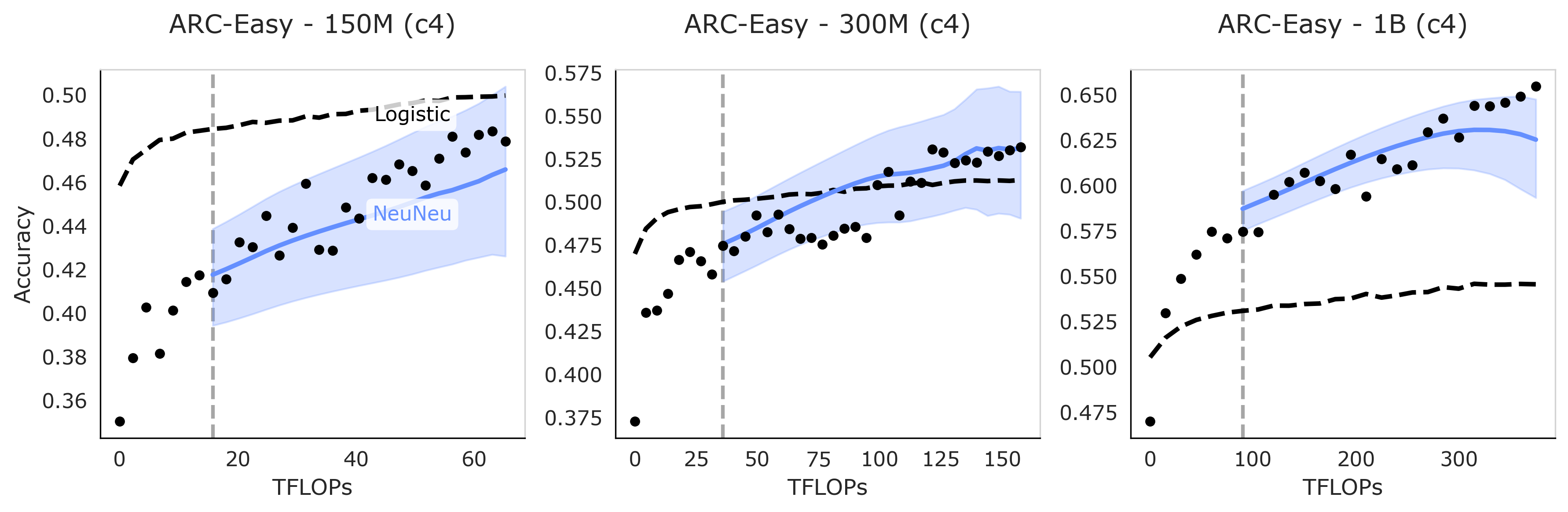}
\caption{Visualizing \method's predictions. Black dots are ground truth accuracies, and the grey line marks the beginning of \method's predictions, after observing the first 20\% of the training run. The light-blue band is the 10\%-90\% interquantile interval predicted by \method~itself. \method tightly captures the ground truth data across model scales, while logistic scaling laws overpredict performance for the 150M model and underpredict performance for the 1B model.}
\label{fig:qualitative}
\end{figure*}

\paragraph{Visualization: ARC-Easy across scales.} 
In Figure \ref{fig:qualitative}, we contrast predictions from \method and \textsc{Logistic} on ARC-Easy, a task representing roughly median predictive performance for \method. Predictions for all other tasks can be found in Appendix Figures \ref{fig:150M} through \ref{fig:1B}. Across the 150M to 1B training runs, \method has tighter fit to the ground truth accuracies and adjusts its predictions based on the model scale. Conversely, we observe that the logistic scaling law overpredicts performance for the 150M model and underpredicts performance for the 1B model.

\subsection{\method Gives Calibrated Uncertainty Estimates}
\label{sec:calibration}

Figure \ref{fig:task_and_quantile}B computes the percentage of ground truth accuracies that land within the neural models' predicted interquantile interval. 74.6\% and 77.6\% of the data lands within the $10\%-90\%$ interquantile interval of \textsc{NoLoss} and \method, suggesting that the interquantile interval predicted by the neural models are close to well-calibrated.
This calibration is useful because scaling predictions are often used to make decisions before a training run has completed. Unlike our parametric baselines, \method{} can indicate when future downstream performance is uncertain, allowing practitioners to distinguish confident extrapolations from cases where additional training may be warranted.

% \begin{figure*}[t]
% \centering
% \includegraphics[width=\textwidth]{figures/combined_ablation_figure.png}
% \caption{A: Neural methods achieve lower error for all extrapolation horizons. B: When increasing the context of observed accuracies $y_{<t}$, \method's prediction error decays exponentially. C: The confidence intervals learned via quantile regression for our neural methods is nearly well-calibrated, containing around 75\% of ground truth accuracies within a $\hat{q}_{0.1}$ to $\hat{q}_{0.9}$ interquantile range.}
% \label{fig:error_and_calibration}
% \end{figure*}

\subsection{\method Reliably Ranks Competing Training Runs}
\label{sec:ranking}

Finally, the most important factor in using \method~over logistic scaling laws would be whether \method~helps make better decisions: given the start of a training run, can we predict whether it will be better than another run with different hyperparameters? 

To study this, we evaluate whether \method~can predict which of two different model configurations will have a better final performance, given the initial 20\% of the training trajectory. For each task $t$ and model $m$, we choose another model $m'$ with a different training configuration, yielding two model-task pairs $(m, t)$ and $(m', t)$. We use all methods to extrapolate the models' performance to the end of training, and the predictor is correct if it correctly ranks the two models---that is, if $\hat{y}_{m,T} > \hat{y}_{m',T}$ matches the true ranking $y_{m,T} > y_{m',T}$. We do this for all possible pairs.

\begin{figure*}[t]
\centering
\includegraphics[width=\textwidth]{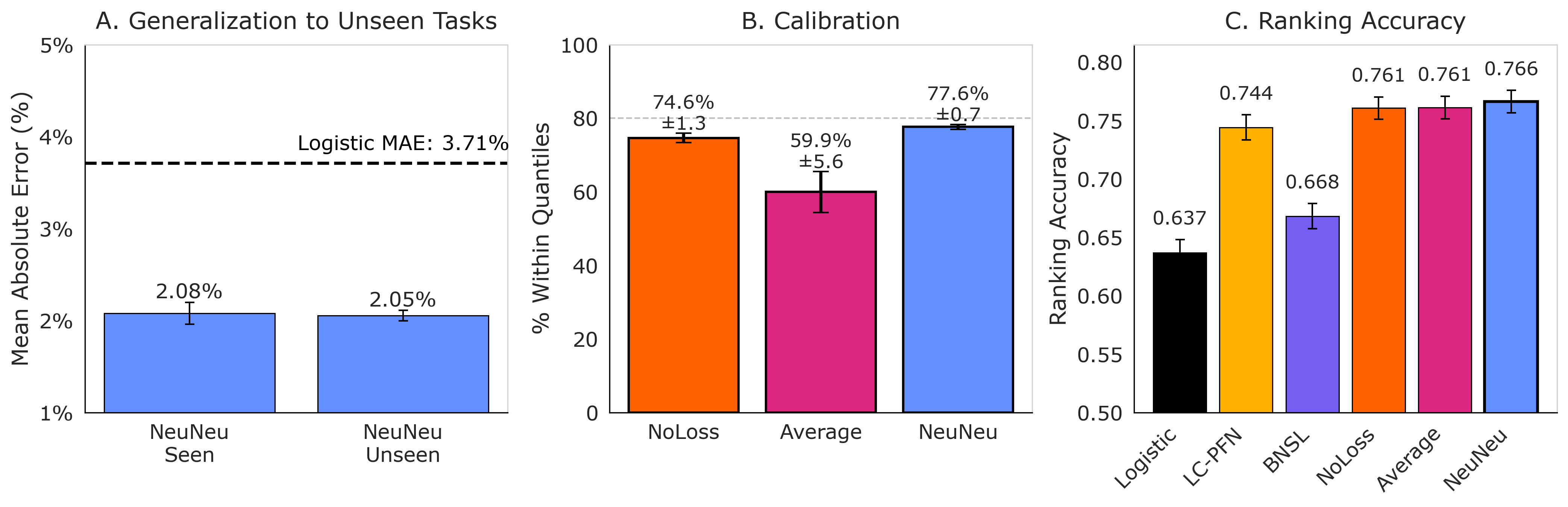}
\caption{
\method{} generalizes beyond the settings seen during meta-training and provides uncertainty estimates that support downstream decision-making.
(A) \method{} maintains low error on downstream tasks held out during training, outperforming logistic scaling laws fit directly to those tasks.
(B) The predicted 10\%--90\% interquantile interval from \method{} contains 77.6\% of ground truth accuracies, close to the 80\% target.
(C) \method{} achieves the highest ranking accuracy when predicting which of two model configurations will reach better final downstream performance.
}
\label{fig:task_and_quantile}
\end{figure*}

In Figure~\ref{fig:task_and_quantile}C, \method~achieves the highest ranking accuracy of 0.766 compared to 0.637 for \textsc{Logistic}, a relative improvement of 20.2\%. These results demonstrate that \method's lower prediction error translates to practical utility: better predictions enable better decisions about which models to train, which hyperparameters to use, and how to allocate limited compute budgets.

\section{Related Work}

% A brief history of scaling laws
Over the decades, a number of works discovered and (rediscovered) power law scaling with respect to data \citep{cortes1993learning, hestness2017deep}. With the advent of language model pretraining \citep{peters-etal-2018-deep, radford2018improving, radford2019language}, data became abundant, and the focus shifted towards scaling compute. \citet{rosenfeld2020a} discovered that loss exhibits power law scaling with respect to parameters as well as data, and soon after \citet{kaplan2020scaling} named this phenomenon \textit{scaling laws}, popularizing the idea by investigating its implications for language models. Later, \citet{hoffmann2022training} refined the method by proposing different ways to estimate scaling laws and recommending the most widely used power law functional form: $L(N, D) = e + \frac{a}{N^\alpha} + \frac{b}{D^\beta}$.
This approach remains the basis for how scaling laws are applied to pretraining today.

% Shortcomings of scaling laws for downstream tasks
Unfortunately, translation from pretraining to downstream tasks has proven more difficult. \citet{wei2022emergent} documented how models display \textit{emergent capabilities}, or capabilities that appear suddenly at scale. Such capabilities are hard to predict, because model performance appears the same at smaller scales. The choice of evaluation metric can ease or exacerbate the problem of emergence \citep{schaeffer2023are, schaeffer2025why}, but even with carefully constructed metrics, extrapolating downstream performance remains a challenge \citep{du2024understanding}. As \citet{liu-etal-2025-just} note, factors beyond compute or loss impact scaling laws; we embrace this fact by providing \method~richer input representations.

A major obstacle for extrapolation is the diversity of scaling behaviors. While many tasks improve with scale, others exhibit \textit{inverse scaling}, where performance gets worse \citep{mckenzie2023inverse, wilcox2024bigger} or does so at first only to become U-shaped \citep{wei2023inversescalingushaped}. To capture these behaviors, researchers have tried creating more complex parametric forms \citep{alabdulmohsin2022revisiting, caballero2023broken}, predicting performance directly from data, parameters, and compute \citep{openai2024gpt4technicalreport, krajewski2025revisitingscalingpropertiesdownstream}, and predicting performance from intermediate task losses such as pretraining loss or the probability of the correct answer \citep{grattafiori2024llama3herdmodels, huang2024compression, gadre2025language, bhagia2025establishing, chen2025scaling}. Despite these efforts, reliably predicting downstream scaling remains a challenge \citep{lourie-etal-2025-scaling}.

% Our work
Our work attempts to move scaling laws beyond parametric assumptions. In doing so, it relates closely to a precursor of scaling laws: trajectory forecasting. Trajectory forecasting has long been studied in hyperparameter optimization; for example, Freeze-Thaw Bayesian Optimization forecasts the asymptotic performance of partially trained models to dynamically allocate compute \citep{swersky2014freezethawbayesianoptimization}. This perspective has recently evolved into meta-learning approaches such as LC-PFN \citep{LC-PFN} and FT-PFN \citep{ft-pfn}, which use transformers trained on synthetic functions to perform in-context inference over learning curves. Our work shares the motivation of these methods but diverges in methodology: rather than using synthetic priors, we learn to extrapolate directly from open-source training runs using quantile regression.

\section{Discussion}
\label{sec:discussion}

\method demonstrates that downstream scaling law prediction benefits from moving beyond parametric forms. Across 66 tasks, \method achieves 1.99\% mean absolute error, a 44\% reduction over logistic scaling laws, while generalizing zero-shot to unseen tasks, model families, and parameter counts over an order of magnitude larger than seen during training. \method also accurately predicts the scaling behavior of OLMo-Hybrid, a non-transformer architecture, despite being trained on only transformer training runs. Beyond pointwise accuracy, \method produces calibrated uncertainty estimates and, most consequentially for practitioners, accurately predicts which training configuration will be better 76.6\% of the time, a 20.2\% relative improvement over logistic baselines.

The ablations in Figure \ref{fig:main}(a) disentangle the sources of \method's improvement over conventional logistic scaling laws. Replacing the parametric form with a neural predictor trained on observed accuracy trajectories accounts for roughly 90\% of the reduction in error (\textsc{Logistic} to \textsc{NoLoss}), while token-level validation losses provide the remaining 10\% (\textsc{NoLoss} to \method). By contrast, average-loss inputs offer no benefit over accuracy trajectories alone. Thus, although model flexibility and accuracy trajectories drive most of \method's gains, token-level losses indeed contain additional predictive signal that is lost through averaging.

% bitter lesson
Our results suggest that the limitations of existing scaling laws are not fundamental to the prediction problem, but rather artifacts of restricted hypothesis classes. Logistic scaling laws embody a strong inductive bias (that downstream performance is a monotonically increasing, saturating function) which is unequipped to deal with tasks exhibiting diverse scaling behaviors. \method~trades this parametric bias for flexibility, learning to recognize patterns in accuracy trajectories and token-level losses rather than assuming a functional form. Just as flexible models of learning curves benefit AutoML \citep{LC-PFN}, we have shown flexible models of language model scaling improve upon scaling laws---beyond what a more general model of learning curves can achieve. Taking a step back, methods that scale with compute and data tend to outperform those relying on human-designed structure \citep{sutton2019bitter}; our work extends this principle to the meta-problem of predicting language model performance itself. As the ecosystem of open-source model checkpoints grows, we expect neural neural scaling laws to improve further.

\paragraph{Toward Foundation Models of Training Dynamics.}
%\emph{World models} learn to simulate environment dynamics, enabling planning and reasoning without direct interaction \citep{ha2018world,bruce2024genie}.
As research moves beyond treating language models as a black box, training dynamics have become an important lens through which to understand them.
\method~can be viewed as a nascent model of training dynamics: given a model's current state (performance) and an implicit action (more compute), it predicts the future state. Costs for training language models are significant and growing \citep{gpt3,morrison2025holistically}; a foundation model that accurately simulates training dynamics could enable practitioners to explore the space of hyperparameters, architectures, and data mixtures without the cost of real experiments. \method and other recent works \citep{hu2023latent,zhang2026configuration} represent steps in this direction.

\paragraph{Limitations and Future Work.} Our approach has several limitations that suggest directions for future work.
We currently consider downstream classification tasks where accuracy is the natural metric, and generative tasks may exhibit different scaling dynamics.
Our experiments are limited to today's open-source checkpoints, but as more training runs become available, future work can train increasingly capable models of training dynamics.
Our quantile regression approach could also be combined with other distribution-free methods like conformal prediction to provide coverage guarantees \citep{Romano2019ConformalizedQR}.
Finally, \method~can serve as a foundation for scaling law theories: by interpreting what features of the loss distribution the CNN encoder extracts, or what patterns in accuracy trajectories the transformer attends to, we may discover better parametric predictors.

\section*{Acknowledgments}

Many thanks to Megan Richards for feedback and comments. MYH and JP are supported by the NSF Graduate Research Fellowship. This work was supported in part through the NYU IT High Performance Computing resources, services, and staff expertise. This work was supported by the Institute of Information \& Communications Technology Planning \& Evaluation (IITP) with a grant funded by the Ministry of Science and ICT (MSIT) of the Republic of Korea in connection with the Global AI Frontier Lab International Collaborative Research. This work was also supported by the Samsung Advanced Institute of Technology (under the project Next Generation Deep Learning: From Pattern Recognition to AI) and the National Science Foundation (under NSF Award 1922658).

\bibliography{example_paper}
\bibliographystyle{plainnat}  

%%%%%%%%%%%%%%%%%%%%%%%%%%%%%%%%%%%%%%%%%%%%%%%%%%%%%%%%%%%%%%%%%%%%%%%%%%%%%%%
%%%%%%%%%%%%%%%%%%%%%%%%%%%%%%%%%%%%%%%%%%%%%%%%%%%%%%%%%%%%%%%%%%%%%%%%%%%%%%%
% APPENDIX
%%%%%%%%%%%%%%%%%%%%%%%%%%%%%%%%%%%%%%%%%%%%%%%%%%%%%%%%%%%%%%%%%%%%%%%%%%%%%%%
%%%%%%%%%%%%%%%%%%%%%%%%%%%%%%%%%%%%%%%%%%%%%%%%%%%%%%%%%%%%%%%%%%%%%%%%%%%%%%%
\newpage
\appendix

\begin{figure}[h!]
  \centering
  \includegraphics[width=\columnwidth]{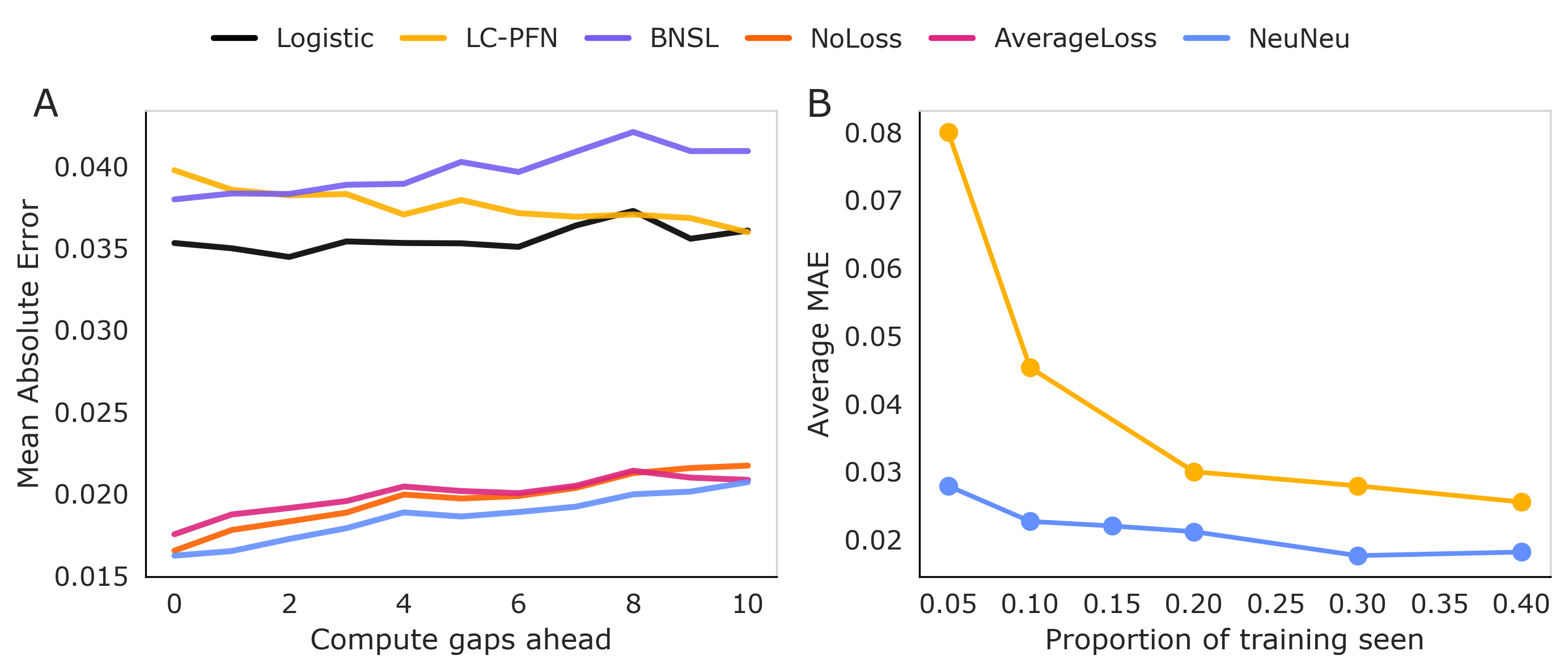}
  \caption{Additional extrapolation results. (A) \method and the neural ablations maintain low error as the extrapolation horizon increases, while parametric and general-purpose learning-curve baselines remain substantially higher. (B) LC-PFN improves as more of the training trajectory is observed, indicating that it is inferring from context.}
  \label{fig:horizon}
\end{figure}

\begin{figure}[h!]
  \centering
  \includegraphics[width=0.8\columnwidth]{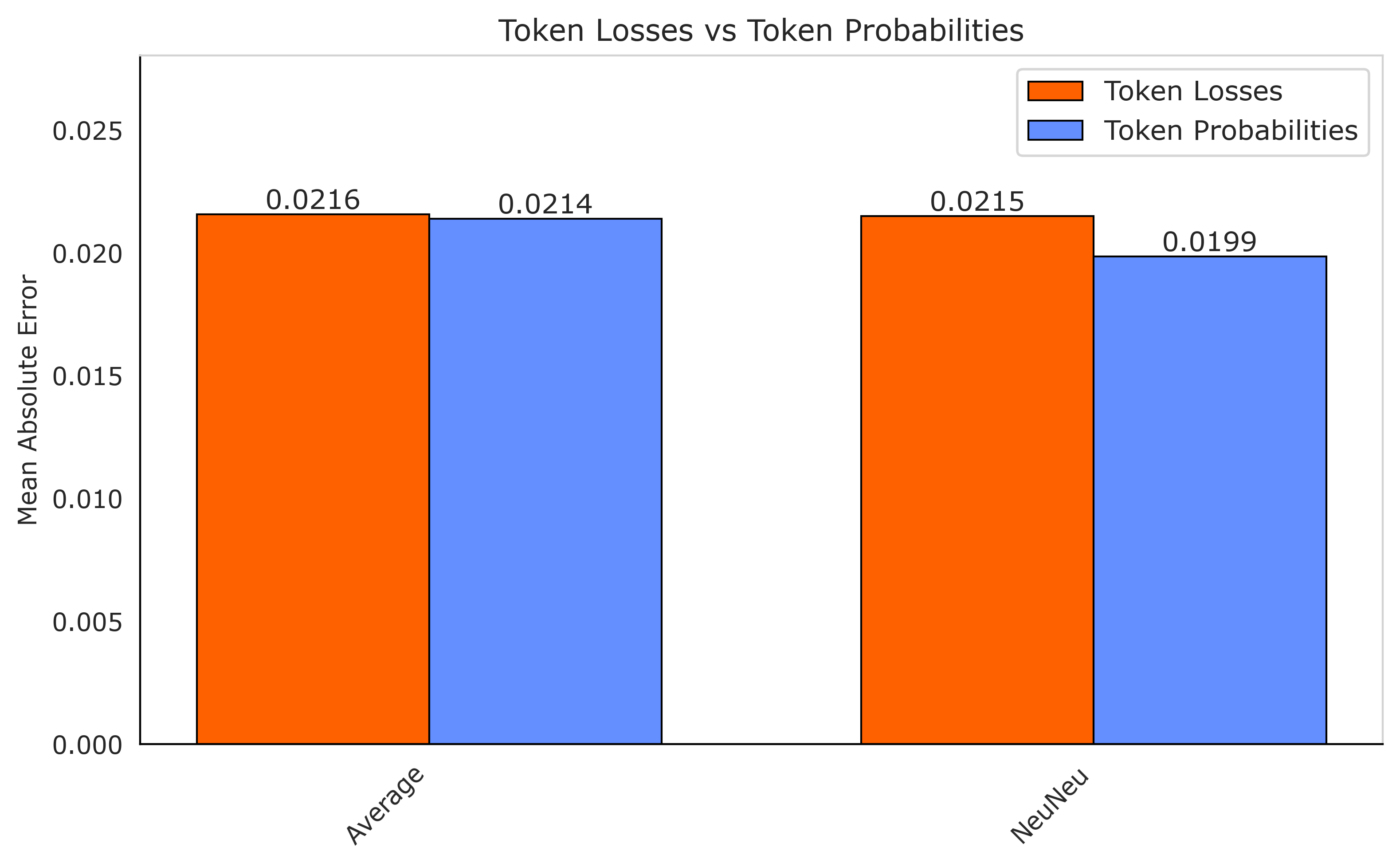}
  \caption{Using token probabilities produces better neural predictors than token losses. Another reason to use probabilities is that the function $e^{-x}$, or the conversion from loss to probability, has larger derivative for smaller loss values, meaning that small changes in loss near convergence for the language model are amplified. This likely matters more than changes when the loss is large and language model performance on downstream tasks is near-chance.}
  \label{fig:prob-ablation}
\end{figure}

\section{Reproducibility}
\label{sec:reproducibility}

\begin{table}[h]
\centering
\caption{Model hyperparameters, shared across our neural models. We chose learning rate and weight decay values based on models of similar size in \cite{magnusson2025datadecide}.}
\label{tab:hyperparameters}
\begin{tabular}{ll}
\toprule
\textbf{Hyperparameter} & \textbf{Value} \\
\midrule
\multicolumn{2}{l}{\textit{Transformer Encoder Architecture}} \\
Hidden dimension & 512 \\
Transformer layers & 6 \\
Attention heads & 8 \\
Feed-forward dimension & 2048 \\
Max sequence length & 512 \\
\multicolumn{2}{l}{\textit{Convolutional Neural Network}} \\
CNN channels & [8, 16, 32, 64] \\
CNN kernel size & 64 \\
CNN stride & 16 \\
Number of groups for GroupNorm & $\min$(8, number of channels) \\
Padding & kernel size $/$ 2 \\
\midrule
\multicolumn{2}{l}{\textit{Training}} \\
Optimizer & AdamW \\
Batch size & 256 \\
Learning rate & $6 \times 10^{-4}$ \\
Weight decay & 0.033 \\
Epochs & 3 \\
Warmup ratio & 0.1 \\
Max gradient norm & 1.0 \\
Drop gap probability (see \S \ref{sec:data}) & 0.4 \\
Random seed & $\{0,1,2,3,4\}$ \\
\midrule
\multicolumn{2}{l}{\textit{Data}} \\
Max encoder tokens & 256{,}000 \\
Quantiles & [0.1, 0.25, 0.5, 0.75, 0.9] \\
\bottomrule
\end{tabular}
\end{table}

From DataDecide \citep{magnusson2025datadecide}, we meta-train on trajectories from the model sizes \{90M, 150M, 300M, 530M, 750M, 1B\} and all pretraining datasets except C4, which we use for evaluation. In total, this yields 6 model sizes $\times$ 24 pretraining datasets = 144 training configurations, where each training configuration contains a sequence of model checkpoints.

To get validation losses for these model checkpoints, we evaluated them on shard 141 from the WebOrganizer dataset \citep{wettig2025organize}, which we chose randomly from all shards and can be accessed at \url{https://huggingface.co/datasets/WebOrganizer/Corpus-200B}. \method takes random token probability spans of length $256{,}000$ as input during training. During evaluation, we take the first span from shard 141 for simplicity.

\paragraph{Compute.} All experiments run on a cluster with a mix of L40S and H200 GPUs. We trained \textsc{NoLoss}, \textsc{Average}, and \method on L40S GPUs for roughly 2 GPU hours. Inference for larger models like Pythia-12B and OLMo-Hybrid were done on H200 GPUs.

We compare our meta-model against two baselines: LC-PFN \citep{LC-PFN}, a learning-curve extrapolator, and Broken Neural Scaling Laws (BNSL) \citep{caballero2023broken}, fit as a zero-shot loss-to-accuracy mapping in the same regime as our logistic baseline. 

\subsection{LC-PFN}

We use the public \texttt{lcpfn} package without retraining. LC-PFN is a prior-data fitted network with fixed weights at release. The model is loaded once per evaluation run via \texttt{LCPFN()} in \texttt{eval} mode.

For each (model, task) trajectory we condition on the first $20\%$ of checkpoints and predict accuracy at every remaining checkpoint. Training steps are normalized to $[0,1]$ by dividing by the maximum step in the \emph{full} trajectory (not the context window) so that target positions on the curve are preserved; accuracies are passed through unchanged. We query \texttt{LCPFN} with
\[
\hat{y}_\text{test} = \texttt{predict\_quantiles}(x_\text{train}, y_\text{train}, x_\text{test}, q),
\]
$q = \{0.1, 0.25, 0.5, 0.75, 0.9\}$. The median is used as the point prediction; the other quantiles supply the predictive intervals reported in the uncertainty plots.

Figure \ref{fig:horizon}B suggests that LC-PFN is working as intended. Like NeuNeu, LC-PFN is a transformer that performs in-context inference, and is not tuned after pretraining. When giving more context to LC-PFN, its prediction error over the remaining accuracies also decreases. Thus, we conclude that LC-PFN is indeed inferring from the existing trajectory, but begins from higher error because it is not specifically trained on the task of downstream scaling prediction.

\subsection{BNSL}

We treat BNSL as a zero-shot scaling-law baseline analogous to the logistic scaling laws. A one-break BNSL curve is fit per task on the training corpus of (average loss, accuracy) pairs and then applied to the eval model's ground-truth average losses to produce predicted accuracies at every checkpoint. No accuracy from the eval trajectories are observed.

\paragraph{Functional form.}
We use the one-break form from \citet{caballero2023broken},
\[
y(x) \;=\; a + b\,x^{-c_0}\bigl(1 + (x/d_1)^{1/f_1}\bigr)^{-c_1 f_1},
\]
fitted to error $y = 1 - \text{acc}$ rather than accuracy, since BNSL models a positive decreasing quantity. Predictions are mapped back via $\hat{\text{acc}} = \text{clip}(1 - \hat{y}, 0, 1)$. For the input $x$, average loss $\ell$ is decreasing, so we transform it to a positive progress measure $x = s / \ell$, where $s$ is the median observed token-level loss. To train BNSL, we follow the curve fitting advice in \citet{caballero2023broken}. BNSL is a deterministic curve fit and does not contribute uncertainty estimates.

\begin{figure}[h!]
  \centering
  \includegraphics[width=\columnwidth]{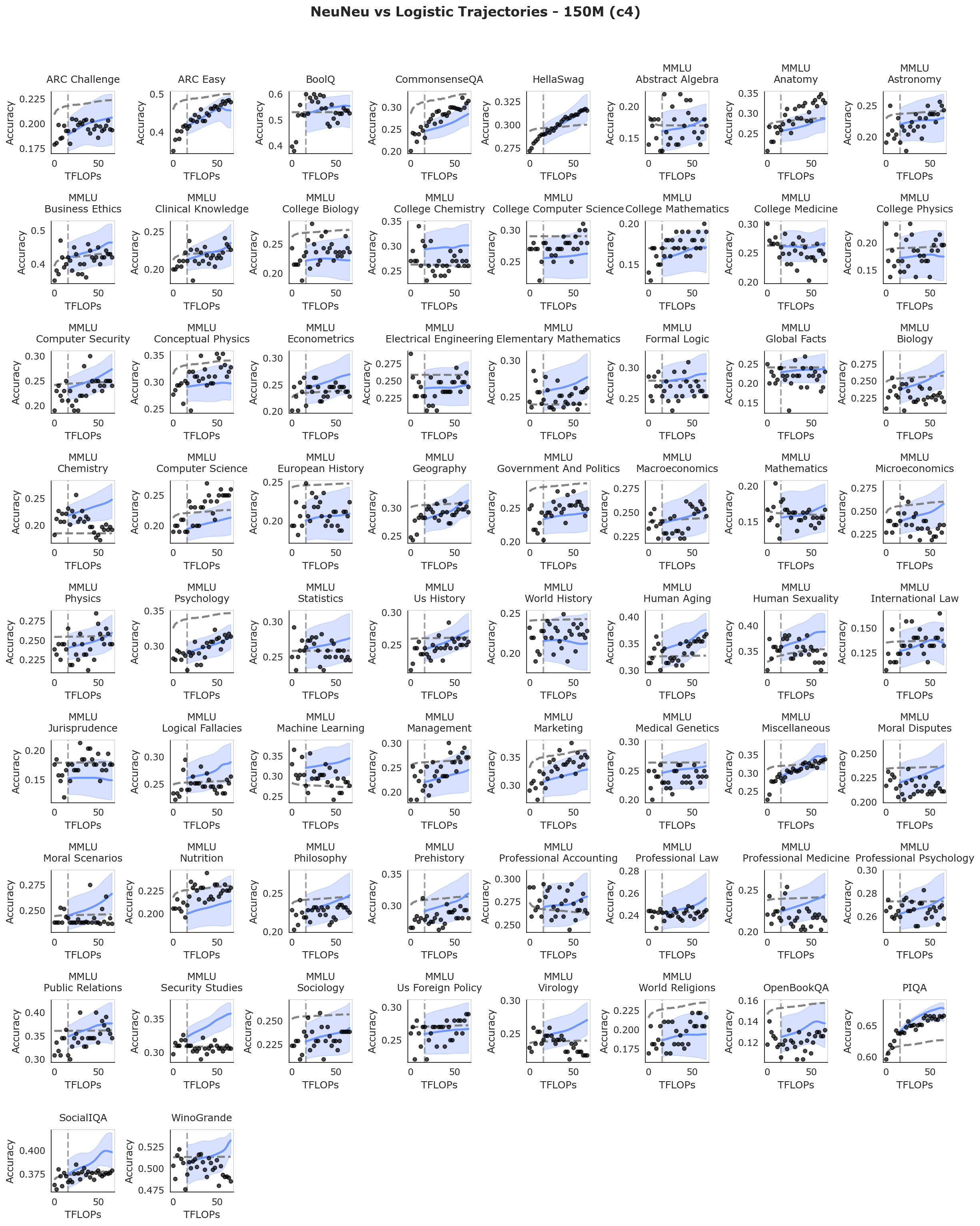}
  \caption{Blue: \method. Dark grey: Logistic scaling law fitted to the task on the training set (\S \ref{sec:data}).}
  \label{fig:150M}
\end{figure}

\begin{figure}[h!]
  \centering
  \includegraphics[width=\columnwidth]{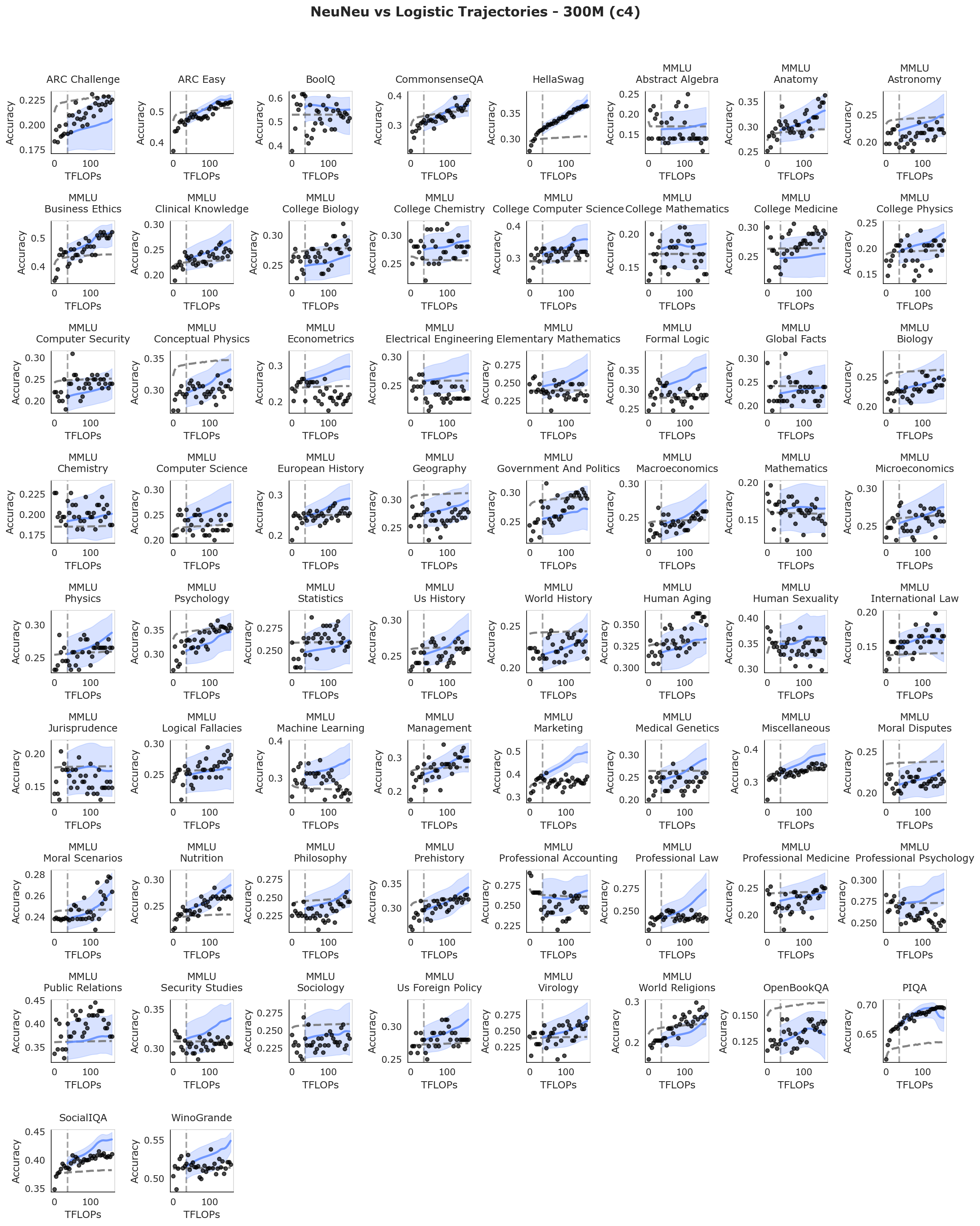}
  \caption{Blue: \method. Dark grey: Logistic scaling law fitted to the task on the training set (\S \ref{sec:data}).}
  \label{fig:300M}
\end{figure}

\begin{figure}[h!]
  \centering
  \includegraphics[width=\columnwidth]{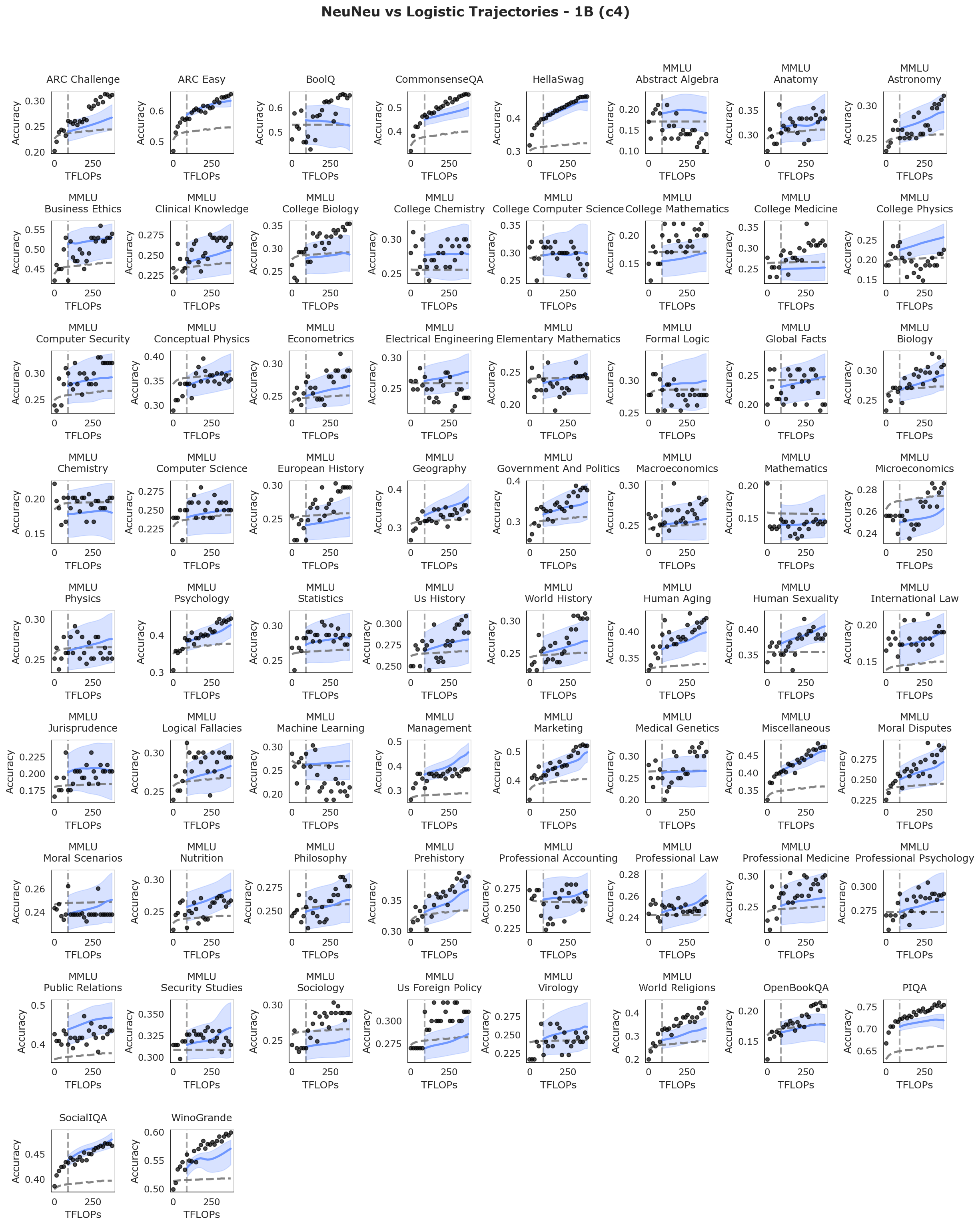}
  \caption{Blue: \method. Dark grey: Logistic scaling law fitted to the task on the training set (\S \ref{sec:data}).}
  \label{fig:1B}
\end{figure}

% \FloatBarrier

% \newpage
% \input{checklist.tex}

\end{document}